\documentclass[]{ceurart}
\usepackage{listings}
\usepackage{siunitx}

\begin{document}

\copyrightyear{2025}
\copyrightclause{Copyright for this paper by its authors.
  Use permitted under Creative Commons License Attribution 4.0
  International (CC BY 4.0).}

\conference{CLEF 2025 Working Notes, 9 -- 12 September 2025, Madrid, Spain}

\title{Enhancing Biomedical Named Entity Recognition using GLiNER-BioMed with Targeted Dictionary-Based Post-processing for BioASQ 2025 task 6}

\author[1]{Ritesh Mehta}[
    email=rmehta307@gatech.edu,
]
\cormark[1]

\address[1]{Georgia Institute of Technology, North Ave NW, Atlanta, GA 30332}
\cortext[1]{Corresponding author.}

\begin{abstract}
Biomedical Named Entity Recognition (BioNER), task6 in BioASQ (A challenge in large-scale biomedical semantic indexing and question answering), is crucial for extracting information from scientific literature but faces hurdles such as distinguishing between similar entity types like genes and chemicals.
This study evaluates the GLiNER-BioMed model on a BioASQ dataset and introduces a targeted dictionary-based post-processing strategy to address common misclassifications.
While this post-processing approach demonstrated notable improvement on our development set, increasing the micro F1-score from a baseline of 0.79 to 0.83, this enhancement did not generalize to the blind test set, where the post-processed model achieved a micro F1-score of 0.77 compared to the baseline's 0.79.
We also discuss insights gained from exploring alternative methodologies, including Conditional Random Fields.
This work highlights the potential of dictionary-based refinement for pre-trained BioNER models but underscores the critical challenge of overfitting to development data and the necessity of ensuring robust generalization for real-world applicability.
\end{abstract}


\maketitle

\section{Introduction}

\subsection{Motivation and Background}
The exponential growth of biomedical literature necessitates advanced text mining techniques to extract valuable information and accelerate scientific discovery.
Biomedical Named Entity Recognition (BioNER) plays a pivotal role in this endeavor by automatically identifying and classifying key entities such as genes, proteins, chemicals, diseases, and anatomical locations within unstructured text.
Accurate BioNER is a fundamental step for various downstream applications, including the construction of knowledge graphs, support for drug discovery pipelines, enhancement of literature search and retrieval systems, and the foundation for complex biomedical question answering.
By structuring the vast amount of information embedded in scientific publications and clinical notes, BioNER facilitates deeper insights and more efficient research workflows.

\subsection{Challenges in BioNER}
Despite its importance, BioNER remains a challenging task due to the inherent complexities of biomedical language.
Entity names often exhibit significant ambiguity and variability, with a single term potentially referring to different entity types depending on context (e.g., a term acting as both a gene and a chemical).
Misclassification between closely related categories, such as genes and chemicals or different types of biomedical techniques, is a common issue.
Furthermore, achieving optimal span segmentation for entities—correctly identifying their precise start and end boundaries—can be difficult, leading to fragmented or overly broad predictions.
The correct identification and normalization of abbreviations and their corresponding long forms also present an ongoing challenge.
These complexities require robust and nuanced approaches to BioNER.

\subsection{The BioASQ Challenge}
Our work is specifically related to Task 6 of the BioASQ \cite{BioASQ2025} CLEF Lab 2025, known as GutBrainIE \cite{BioASQ2025taskGutBrainIE} CLEF 2025, an NLP challenge supported by the EU project HEREDITARY.
The goal of the GutBrainIE challenge is to foster the development of Information Extraction (IE) systems capable of extracting structured information from biomedical abstracts.
The focus is particularly on literature related to gut microbiota and its intricate connections with Parkinson's disease and mental health, thereby aiming to support expert understanding of the complex gut-brain interplay.
The GutBrainIE challenge itself is divided into two main tasks: Named Entity Recognition (NER) and Relation Extraction.
Our research presented in this paper specifically addresses Subtask 6.1 - Named Entity Recognition.
In this subtask, participants are tasked with identifying and classifying specific text spans from PubMed abstracts, which are focused on the gut-brain axis, into one of 13 predefined categories, including common biomedical entities such as bacteria, chemical, and microbiota.
The challenge provides a diverse set of training data, comprising highly curated "Gold" and "Platinum" collections annotated by domain experts, "Silver" collections annotated by trained students, and a "Bronze" collection generated through distant supervision.
This rich resource is designed for system development, with the test set being derived from the Gold and Platinum collections to ensure high quality evaluation.

\subsection{Paper Structure}
This paper details our investigation into enhancing BioNER performance, focusing on the use of a pre-trained model, GLiNER-BioMed, augmented with targeted post-processing techniques.
Section 2 reviews related work in the field of BioNER.
Section 3 describes our methodology, including a detailed overview of the dataset used, the GLiNER-BioMed pipeline implemented, and the development of our dictionary-based post-processing rules.
Section 4 presents our experimental setup, the baseline results obtained, the measured impact of our post-processing methods, and an analysis of alternative approaches that were explored during the course of this research.
Following this, Section 5 discusses these results, interpreting the effectiveness of our methods, analyzing persistent errors, and acknowledging the limitations of the study.
Finally, Section 6 outlines potential avenues for future work, and Section 7 concludes the paper with a summary of our contributions and key findings.

\section{Related Work}

The field of BioNER has made substantial progress, moving from early dictionary-based and rule-based systems to more sophisticated machine learning and deep learning approaches. This section reviews literature pertinent to the methods and challenges addressed in our study.

\subsection{Pre-trained Language Models in BioNER}
The emergence of transformer-based pre-trained language models has revolutionized the field of biomedical named entity recognition, providing substantial improvements over traditional approaches. 
The introduction of BERT (Bidirectional Encoder Representations from Transformers) \cite{Devlin2019BERTPO} marked a significant milestone in natural language processing, leading to the development of several domain-specific variants tailored for biomedical text mining.

BioBERT represents one of the most influential developments in biomedical NLP, being the first domain-specific BERT-based model pre-trained on large-scale biomedical corpora \cite{Lee2019BioBERTAP}.
It demonstrated that pre-training BERT on biomedical corpora (PubMed abstracts and PMC full-text articles) substantially improves performance on biomedical tasks, achieving a 0.62\% F1 score improvement in biomedical NER compared to the original BERT model.
The model was trained for 23 days on eight NVIDIA V100 GPUs and has shown consistent superior performance across various biomedical text mining tasks including NER, relation extraction, and question answering.

Following BioBERT's success, several other domain-specific models have been developed. PubMedBERT, specifically trained on PubMed abstracts, has shown excellent performance in clinical trial protocols and biomedical NER tasks \cite{Li2022ACS} .
Comparative studies have demonstrated that PubMedBERT achieves F1-scores of 0.715, 0.836, and 0.622 on different clinical trial eligibility criteria corpora, outperforming general-domain transformer models.
SciBERT \cite{Beltagy2019SciBERTAP}, another notable variant, was pre-trained on scientific text and has achieved state-of-the-art results on several biomedical datasets including BC5CDR and ChemProt.
ClinicalBERT \cite{Huang2019ClinicalBERTMC}, specialized for clinical tasks, has proven highly effective for healthcare applications requiring understanding of clinical narratives.

More recently, GLiNER-biomed \cite{Yazdani2025GLiNERbiomedAS} has introduced a novel approach to biomedical NER by enabling zero-shot entity recognition through natural language descriptions.
Unlike conventional approaches that rely on fixed taxonomies, GLiNER uses natural language descriptions to infer arbitrary entity types, achieving a 5.96\% improvement in F1-score over the strongest baseline in zero- and few-shot scenarios.
This approach addresses the challenge of recognizing novel entities without requiring extensive retraining, making it particularly valuable for the rapidly evolving biomedical domain.

The effectiveness of these pre-trained models stems from their ability to capture complex biomedical language patterns and domain-specific knowledge through large-scale unsupervised pre-training.
Specialized transformers-based models encode substantial biological knowledge, although some of this knowledge may be lost during fine-tuning for specific tasks.
The success of these domain-adapted models has established pre-trained transformers as the foundation for state-of-the-art biomedical NER systems.

\subsection{Post-processing Techniques in BioNER}
Post-processing techniques play a crucial role in refining the output of BioNER systems, addressing common issues such as entity boundary detection, label disambiguation, and normalization.
These techniques are particularly important in the biomedical domain, where entity names exhibit significant variability and ambiguity, often requiring additional processing steps to achieve optimal performance.

Dictionary-based post-processing represents one of the most straightforward and effective approaches for improving NER performance.
Simple dictionary-based methods, despite their apparent simplicity, can achieve reasonable performance on well-known NER datasets \cite{Yuchenlin}.
For instance, dictionary-based approaches have achieved F1-scores of 58.15 on CoNLL 2003 English NER and 46.90 on OntoNotes 5.0 NER tasks.
In the biomedical domain, dictionary-based correction is particularly valuable for addressing systematic misclassifications between related entity types, such as the frequent confusion between genes and chemicals.

Rule-based post-processing methods continue to play an important role in BioNER systems, particularly for handling domain-specific patterns and constraints.
These methods can include regular expression-based corrections, boundary adjustment rules, and constraint-based validation of entity predictions \cite{Rais2014ACS}.
While rule-based systems may suffer from limited recall and scalability issues, they often achieve high precision for well-defined patterns and can be particularly effective when combined with machine learning approaches .

Recent work has also explored the integration of post-processing with large-scale biomedical applications.
The BioEN package demonstrates a three-phased approach that includes data parsing, NER enhancement, and model serving for real-time applications \cite{Raza2022LargescaleAO}.
This work highlights the importance of considering post-processing not just as an isolated step, but as an integral component of end-to-end biomedical information extraction systems.

\subsection{Conditional Random Fields for BioNER}
Conditional Random Fields (CRFs) have historically played a fundamental role in biomedical named entity recognition, serving as both standalone models and as components in hybrid neural architectures.
Prior to the deep learning era, CRFs with rich feature sets were among the most popular and consistently well-performing methods for BioNER tasks \cite{He2008BiologicalER}.
Their ability to model label dependencies and incorporate diverse linguistic features made them particularly well-suited to the complex naming conventions and structural patterns found in biomedical text.

Rahman et al \cite{Hahn2016DiseaseNE} demonstrated the effectiveness of comprehensive feature engineering for disease NER using CRFs on the NCBI disease corpus, achieving maximum F-scores of 94\% for training data using 10-fold cross-validation.
Their approach incorporated sentence and token-level features including orthographic, contextual, affixes, n-grams, part-of-speech tags, and word normalization.

Recent developments have integrated CRFs with advanced pre-trained language models, creating sophisticated architectures such as BERT-BiLSTM-CRF.
These models combine the contextual understanding of transformer-based models with the structured prediction advantages of CRFs \cite{Greenberg2018MarginalLT}.
The BioBBC model exemplifies this trend, utilizing multi-feature embeddings including BERT embeddings, part-of-speech tags, character-level features, and data-specific embeddings within a BERT-BiLSTM-CRF framework \cite{Wang2023NamedER}.
Such approaches have achieved significant improvements on benchmark BioNER datasets by effectively combining the strengths of pre-trained language models with traditional structured prediction methods.

Despite the dominance of transformer-based models in recent years, CRFs continue to play an important role in biomedical NER, particularly in scenarios where interpretability, computational efficiency, or structured prediction constraints are important considerations.
The explicit modeling of label dependencies provided by CRFs remains valuable for ensuring coherent entity predictions, especially in complex biomedical texts where entity boundaries and relationships are crucial for accurate extraction.

\section{Methodology}

This section outlines the systematic approach undertaken in our study, detailing the dataset utilized, the application of the GLiNER-BioMed model, the development and implementation of our dictionary-based post-processing rules, and the metrics used for evaluation.

\subsection{Dataset}
Our research utilized the dataset provided for Subtask 6.1 (Named Entity Recognition) of the GutBrainIE CLEF 2025 challenge.
This dataset is specifically curated to address the complexities of NER within the domain of gut-brain axis research, focusing on PubMed abstracts related to gut microbiota, Parkinson's disease, and mental health.
The challenge provides training data across four tiers of curation: "Platinum" (111 documents, 3,638 entities), "Gold" (208 documents, 5,192 entities), "Silver" (499 documents, 15,275 entities), and "Bronze" (749 documents, 21,357 entities).
For system development and initial evaluation, we leveraged the "Platinum," "Gold," and "Silver" collections, aiming to incorporate a broader range of annotated examples while acknowledging the varying levels of curation.
The "Platinum" collection represents a subset of "Gold" further validated by biomedical experts, offering the highest annotation quality.
The "Silver" collection, annotated by trained students, provided additional volume.
The "Bronze" collection, being distantly supervised without manual revision, was excluded from our training and development pipeline.
The official "Development Set" provided by the challenge consists of 40 documents, containing 1,117 entities with an average of 27.93 entities per document.

The task requires the identification and classification of entity mentions into 13 predefined categories: bacteria, biomedical\_technique, chemical, DDF (Disease, Disorder, or Finding), dietary\_supplement, drug, food, gene, human, animal, anatomical\_location, microbiome, and statistical\_technique.
Entity mentions in the dataset are provided as tuples including the entity label, its location (title or abstract), and character start and end offsets.
In the Development Set, the most frequent entity type is DDF (379 instances, 33.9\% of entities), followed by chemical (131 instances, 11.7\%) and microbiome (127 instances, 11.4\%).
Less frequent but critical categories include gene (39 instances, 3.5\%) and drug (60 instances, 5.4\%).
The full distribution of entity types in the development set is presented in Figure 1.

\begin{figure}[h]
\begin{center}
   \includegraphics[width=1\linewidth]{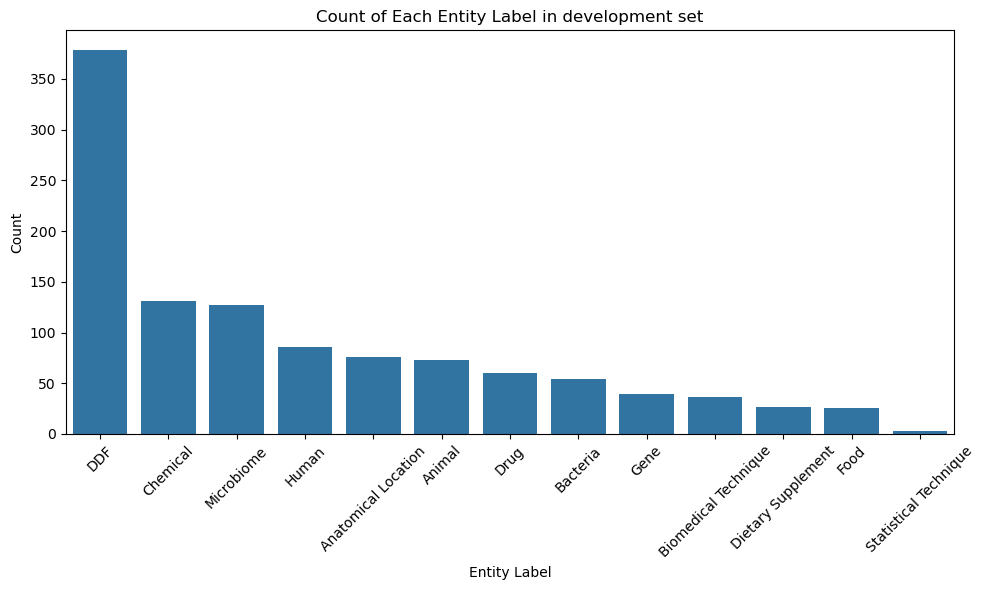}
\end{center}
   \caption{Entity label count in development set}
\label{fig:long-dev_set_count_plot}
\label{fig:onecol-dev_set_count_plot}
\end{figure}

\subsection{GLiNER-BioMed Model Pipeline}
The core of our named entity recognition pipeline employed the GLiNER-BioMed model, a variant of the GLiNER (Generalist model for NER) model specifically adapted for the biomedical domain.
GLiNER is a pre-trained transformer-based model designed for flexible named entity recognition, capable of identifying entities based on provided class labels, often without extensive task-specific fine-tuning.
We utilized the GLiNER-BioMed model as available from the Hugging Face model hub. Initial attempts to use the default tokenizer provided with the model encountered compatibility issues, necessitating a minor workaround.

To adapt the model more closely to the specific nuances of the GutBrainIE dataset and its 13 entity types, the GLiNER-BioMed model was fine-tuned.
The fine-tuning process utilized a combined training set comprising the "Platinum," "Gold," and "Silver" collections from the GutBrainIE challenge data.
The model was trained for 10 epochs, with evaluation performed after every epoch. A training batch size of 4 was used, and the maximum sentence length was set to 384 tokens.
The learning rate for the encoder was 1e-5, while other layers used a learning rate of 5e-5, with a warmup ratio of 0.1.
Token representations were not frozen during training and random dropping was enabled.
Training was conducted on a CUDA-enabled GPU.
For inference on the development and test sets, this fine-tuned GLiNER-BioMed model was provided with the combined text (title and abstract of each document) and the list of 13 target entity labels.

An initial qualitative and quantitative analysis of the output from this fine-tuned GLiNER-BioMed model on our development set revealed several characteristic error patterns.
While the model demonstrated strong overall performance in identifying relevant spans, common issues persisted, including: (i) misclassifications between semantically similar or related entity types, with the most frequent confusion observed between 'gene' and 'chemical' entities; (ii) the prediction of overly generic terms as specific entities; and (iii) occasional suboptimal span boundaries. These are shown in Figure 2.
This initial error analysis called for the development of our targeted post-processing steps.
\begin{figure}[h]
\begin{center}
   \includegraphics[width=0.6\linewidth]{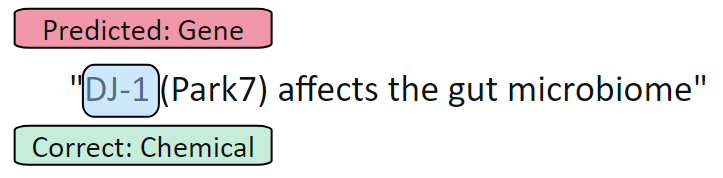}
\end{center}
   \caption{Misclassifications between semantically similar or related entity types}
\label{fig:long-error1}
\label{fig:onecol-error1}
\end{figure}

\begin{figure}[h]
\begin{center}
   \includegraphics[width=0.6\linewidth]{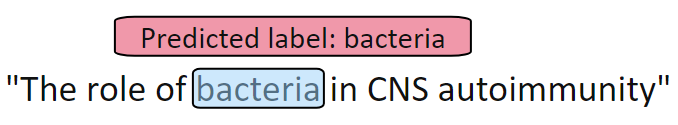}
\end{center}
   \caption{Prediction of overly generic terms as specific entities}
\label{fig:long-error2}
\label{fig:onecol-error2}
\end{figure}

\begin{figure}[h]
\begin{center}
   \includegraphics[width=1\linewidth]{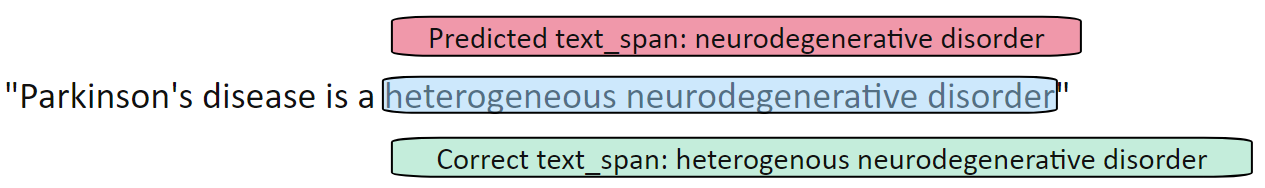}
\end{center}
   \caption{Suboptimal span boundaries}
\label{fig:long-error3}
\label{fig:onecol-error3}
\end{figure}

\subsection{Dictionary-Based Post-processing}
To address the identified error patterns in the GLiNER-BioMed output, we developed and applied a sequence of dictionary-based post-processing rules.
This strategy aimed to leverage curated domain knowledge to refine the model's predictions.

The first rule implemented was the removal of trivial entities.
This rule targeted instances where the identified text\_span was identical (case-insensitively) to its predicted entity\_label.
For example, if the model predicted the literal word "bacteria" as an entity of type 'bacteria', this prediction was removed, as such instances are typically not considered valid entity mentions in this context and often represent false positives.

The second, and more impactful, rule focused on dictionary-based label correction, primarily targeting the frequent confusion between 'gene' and 'chemical' entities. To implement this, we constructed four key dictionaries:
\begin{itemize}
\item \texttt{KNOWN\_GENES}: This dictionary was populated with common gene and protein names, including cytokines (e.g., "IL-6", "TNF-$\alpha$"), growth factors, specific immunoglobulins ("secretory IgA"), transporters ("dopamine transporter"), and other terms frequently annotated as 'gene' in biomedical literature, drawing from the error analysis of the development set.
\item \texttt{KNOWN\_CHEMICALS}: This dictionary contained names of specific chemicals that were sometimes misclassified as genes, such as "DJ-1" or "saline", as well as terms like "curli" (a bacterial amyloid protein often treated as a chemical in context).
\item \texttt{KNOWN\_FOOD}: Included terms such as "nonnutritive sweeteners" (and its abbreviation "NNSs") and "low-carbohydrate high-fat diets" ("LCHF"), which were confused with 'dietary supplement'.
\item \texttt{KNOWN\_DIETARY\_SUPPLEMENTS}: This dictionary helped to correctly classify specific supplement terms like "NS9" that might otherwise be mislabeled.
\end{itemize}

The post-processing logic iterates through GLiNER's predictions. If an entity was predicted as 'chemical' but its text\_span (normalized to lowercase) matched an entry in KNOWN\_GENES, its label was corrected to 'gene'. 
Conversely, if an entity was predicted as 'gene' and its text\_span matched an entry in KNOWN\_CHEMICALS, its label was corrected to 'chemical'.
Similar corrective logic was applied for the KNOWN\_FOOD and KNOWN\_DIETARY\_SUPPLEMENTS dictionaries.

\subsection{Offset and Tag Adjustments}

After all correction rules were applied, each processed entity was assigned a final 'tag' indicating its origin. This tag was set to 't' if the entity's start offset (within the combined\_text) fell within the span of the original title, and 'a' if it fell within the abstract. The model-specific confidence scores was removed from the final entity representation to match the required annotation format.
The output for each document thus consisted of the original metadata (title, abstract, etc.) and a list of processed pred\_entities, each with its text\_span, corrected entity\_label, adjusted start\_idx, end\_idx, and the 't'/'a' tag.

To ensure consistency and adherence to common NER output formats, final adjustments were made to entity offsets.
The GutBrainIE dataset provides entities with offsets relative to their original location (either title or abstract). To adhere to GLiNER input format and analysis, the title and abstract of each document were concatenated into a combined\_text.
Consequently, the character offsets of entities originating from the abstract needed to be adjusted by adding the length of the title.

\subsection{Evaluation Metrics}
The performance of our system, encompassing both the baseline fine-tuned GLiNER-BioMed model and its post-processed versions, was evaluated using the standard metrics of precision, recall, and F1-score.
For a comprehensive evaluation, we report both micro-averaged and macro-averaged scores.
A True Positive (TP) was defined as a predicted entity that exactly matched a ground truth entity in both character span and label.
False Positives (FP) were predicted entities not present in the ground truth, and False Negatives (FN) were ground truth entities missed by the system.

The micro-average computes metrics globally by summing all individual TPs, FPs, and FNs across all 13 entity types before calculating precision, recall, and F1-score.
This approach gives more weight to performance on more frequent entity categories.
Given that the official leaderboard for the Subtask 6.1 was based on the micro-averaged F1-score, this metric served as the primary indicator of overall system performance in our experiments.

The macro-average calculates precision, recall, and F1-score for each of the 13 entity types independently and then computes the unweighted average of these scores.
This method treats all entity categories equally, regardless of their frequency in the dataset, providing insight into the system's consistency across different types of entities.
While the micro-F1 score was the primary focus for leaderboard ranking, macro-averaged scores were also analyzed to understand the system's performance on less frequent but potentially important entity types.

\section{Experiments and Results}
This section details the experimental framework, presents the performance of our baseline and post-processed systems on the Subtask 6.1 development and test sets, analyzes the impact of our post-processing strategies, and discusses learnings from alternative approaches that were explored.

\subsection{Experimental Setup}
All experiments were conducted using Python version 3.10.
The primary natural language processing tasks utilized the Transformers library from Hugging Face for model access and PyTorch as the deep learning framework.
The NLTK library was employed for Part-of-Speech (POS) tagging during the exploration of Conditional Random Field model as detailed in section 4.5.1.
The GLiNER-BioMed model was used as the base NER system.
Fine-tuning of this model was performed on an NVIDIA Quadro M4000 GPU with 8 GB of memory for 10 epochs, using the hyperparameters detailed in Section 3.2.
Post-processing scripts, including dictionary lookups and rule applications, were implemented in Python.
Performance on the development set was measured using the official GutBrainIE evaluation script, and final system performance was determined by submission to the official test set leaderboard.

\subsection{GLiNER-BioMed Baseline Performance}
The fine-tuned GLiNER-BioMed model, without any subsequent post-processing, served as our baseline for development.
On the official GutBrainIE Development Set, this baseline system achieved a micro-averaged F1-score of 0.7857 (Precision: 0.7390, Recall: 0.8389).
The macro-averaged F1-score was 0.6674 (Macro-Precision: 0.6183, Macro-Recall: 0.7727), indicating reasonable performance across most entity types, with some variability for less frequent classes.
The entire system pipeline is shown in Figure 5.

\begin{figure}[hthp]
    \centering
   \includegraphics[width=0.8\linewidth]{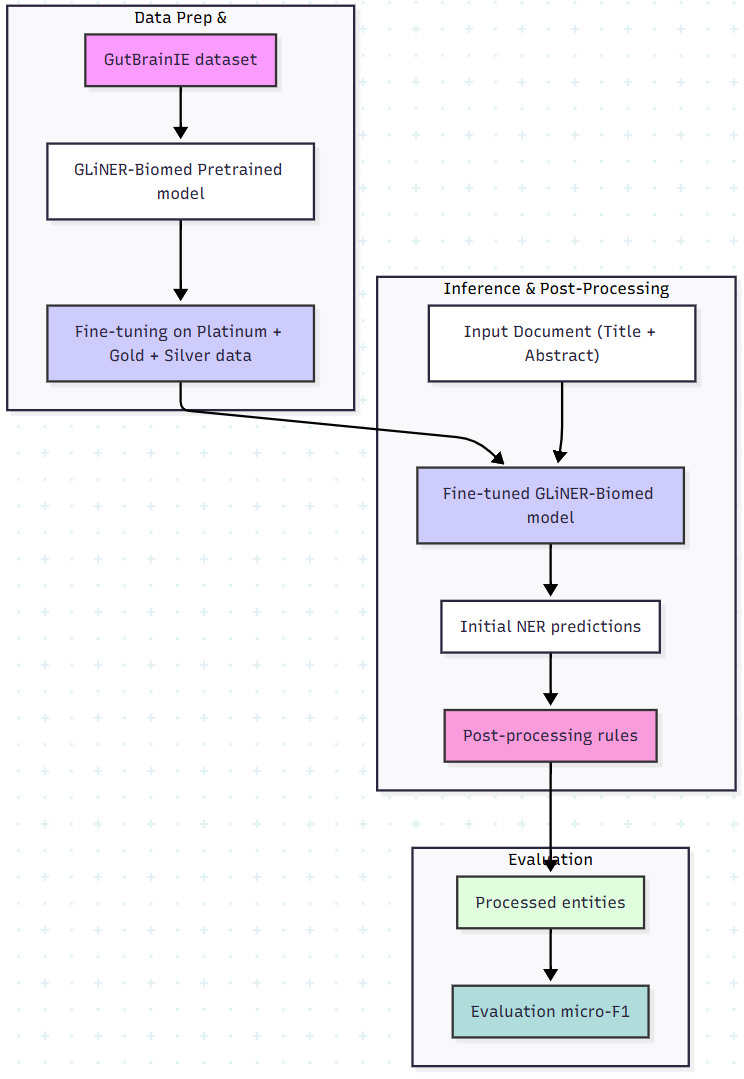}
   \caption{Flowchart of the BioNER system pipeline}
\label{fig:long-system}
\label{fig:onecol-system}
\end{figure}

\subsection{Impact of Dictionary-Based Post-processing}
Following the baseline evaluation, our dictionary-based post-processing rules (as described in Section 3.3) were applied to the output of the fine-tuned GLiNER-BioMed model.
On the Development Set, this post-processing strategy led to an improvement in the primary evaluation metric, with the micro-averaged F1-score increasing from 0.7857 to 0.8316.
This improvement was primarily driven by an increase in micro-precision from 0.7390 to 0.7847 and micro-recall from 0.8389 to 0.8845, suggesting a better balance of correctly identified entities and fewer false positives due to label corrections.
The macro-averaged F1-score also improved from 0.6674 to 0.7766 (Macro-Precision: from 0.6183 to 0.7148, Macro-Recall: from 0.7727 to 0.8805).
Analysis of the specific rules indicated that the dictionary-based label correction for gene/chemical entities contributed most significantly to this observed improvement on the development data.

\subsection{Performance on official test set}
Our final submitted system for the GutBrainIE Subtask 6.1 incorporated the fine-tuned GLiNER-BioMed model followed by the dictionary-based post-processing rules.
On the official blind test set, this system achieved a micro-averaged F1-score of 0.7743 (Micro-Precision: 0.7337, Micro-Recall: 0.8197).
The macro-averaged F1-score on the test set was 0.6872 (Macro-Precision: 0.6342, Macro-Recall: 0.7849). 
Notably, the test set micro F1-score of 0.7743 was lower than the 0.8316 achieved by the same post-processed system on the development set.
Furthermore, it was also slightly lower than the 0.7857 micro F1-score achieved by the baseline GLiNER-BioMed model (without post-processing) when evaluated on the development set.
This discrepancy indicates that the improvements observed from post-processing on the development data did not generalize to the unseen test data; instead, the post-processing step appears to have had a detrimental effect on the final test set performance when compared to its own development set performance and the development set baseline.
Table 1 summarized all the results.

\begin{table}[htbp]
  \centering
  \small 
  \setlength{\tabcolsep}{3pt} 
  \caption{Performance Comparison on Development and Test Sets: Baseline GLiNER-BioMed vs. Post-processed GLiNER-BioMed}
  \label{tab:ner_results_comparison_dev_test}
  \sisetup{} 
  \begin{tabular}{
    >{\raggedright\arraybackslash}p{5.2cm} 
    l 
    S 
    S 
    S 
    S 
    S 
    S 
  }
    \toprule
    \textbf{System Configuration} & \textbf{Dataset} & \multicolumn{3}{c}{\textbf{Micro-Averaged}} & \multicolumn{3}{c}{\textbf{Macro-Averaged}} \\
    \cmidrule(lr){3-5} \cmidrule(lr){6-8}
    & & \multicolumn{1}{c}{Precision} & \multicolumn{1}{c}{Recall} & \multicolumn{1}{c}{F1-score} & \multicolumn{1}{c}{Precision} & \multicolumn{1}{c}{Recall} & \multicolumn{1}{c}{F1-score} \\
    \midrule
    GLiNER-BioMed (Fine-tuned Baseline) & Dev Set & 0.7390 & 0.8389 & 0.7857 & 0.6183 & 0.7727 & 0.6674 \\
    \midrule
    \multirow{2}{5.2cm}{\raggedright GLiNER-BioMed (Fine-tuned + Post-processing)} & Dev Set & 0.7847 & 0.8845 & 0.8316 & 0.7148 & 0.8805 & 0.7766 \\
    & Test Set & 0.7337 & 0.8197 & 0.7743 & 0.6342 & 0.7849 & 0.6872 \\
    \bottomrule
  \end{tabular}
\end{table}

\subsection{Explored Alternative Approaches (Learnings)}
During the course of this research, several alternative methodologies were explored before settling on the dictionary-based post-processing of GLiNER-BioMed.
\subsubsection{Conditional Random Field (CRF) Model}
An attempt was made to build a BioNER system using a Conditional Random Field model.
This involved tokenizing the text and extracting a range of features for each token.
The feature set included basic word-level properties (e.g., the lowercase word, suffixes of length 2 and 3, booleans for all-uppercase, title case, and all-digits), Part-of-Speech (POS) tags (both the full tag and its first two characters), and contextual features from the immediately preceding and succeeding words (including their lowercase form, case properties, and POS tags).
Additionally, flags indicating the beginning (BOS) or end (EOS) of a sentence were incorporated.
This CRF model was trained on the combined Platinum, Gold, and Silver training data.
However, its performance on the development set yielded a micro F1-score which was substantially lower than the GLiNER-BioMed baseline.
Given the significant effort required for traditional feature engineering and the superior performance of the transformer-based approach, this line of investigation was not pursued further for the final system.

\subsubsection{Other Attempted Post-processing Rules for GLiNER}
Beyond dictionary-based correction, other rule-based post-processing strategies were tried.
This included regex-based patterns for label correction (which proved difficult to make both comprehensive and precise, often being too restrictive or overly general) and rules for merging adjacent entity fragments based on connecting POS tags. The entity merging attempts did not yield consistent improvements and sometimes incorrectly combined distinct entities, leading to a decrease in F1-score on the development set.
These explorations highlighted the difficulty in crafting generalizable rule-based systems and reinforced the decision to focus on more targeted dictionary lookups for the most common and clear-cut error patterns.

\section{Discussion}
The results of our experiments provide several key insights into the application of the GLiNER-BioMed model and the utility of dictionary-based post-processing for the Subtask 6.1 on NER.
Our primary approach, combining a fine-tuned GLiNER-BioMed model with targeted dictionary rules, demonstrated notable success on the development set but faced challenges in generalizing to the unseen test set.

\subsection{Effectiveness of Dictionary-Based Post-processing}
The dictionary-based post-processing rules proved effective in improving the micro F1-score on the development set from 0.79 to 0.83.
This improvement can be attributed primarily to the successful correction of systematic misclassifications made by the GLiNER-BioMed model, particularly the confusion between 'gene' and 'chemical' entities.
For instance, terms like "IL-6" or "TNF-$\alpha$", frequently mislabeled as 'chemical' by the baseline model, were correctly identified as 'gene' by our dictionary lookup.
Similarly, terms like "DJ-1", mislabeled as 'gene', were corrected to 'chemical'. The rule for removing trivial entities (where the text span matched the entity label) also contributed by eliminating a class of clear false positives, which likely had a positive impact on precision.
The curated nature of these dictionaries, built upon an analysis of errors on the development set itself, allowed for precise targeting of these common issues.

\subsection{Analysis of Persisting Errors}
A detailed error analysis was conducted on the development set outputs to understand the impact of the post-processing rules.
On this set, where ground truth was available, the dictionary rules successfully corrected numerous instances of 'gene'/'chemical' confusion as intended, contributing to the observed F1-score increase.
For example, terms like "IL-6" initially misclassified as 'chemical' were correctly relabeled as 'gene'.
The removal of trivial entities (where text span matched the label) also helped reduce some false positives.

However, the drop in performance on the blind test set (where the post-processed system scored 0.77, compared to a development set baseline of 0.79 for the raw model) suggests that the overall system did not generalize well. 
Without access to the ground truth for the test set, a direct categorization of errors (False Positives, False Negatives, Incorrect Labels) on this set is not possible.
Nonetheless, the performance degradation implies that one or more of the following occurred: (a) dictionary rules incorrectly changed predictions that would have been correct (or were less incorrect) from the baseline GLiNER-BioMed model on the test set, effectively introducing new errors; (b) the types of errors made by the baseline GLiNER-BioMed model on the test set were different from those on the development set, and our rules were not well-suited to address them, or even exacerbated them; (c) the dictionaries, while curated based on development set errors, were not comprehensive or general enough for the diversity of the test set, potentially misapplying corrections.
The primary challenge highlighted is the difficulty in creating post-processing rules that generalize well from development data to unseen test data, especially when these rules are designed to correct specific error patterns observed in a limited dataset.

\subsection{Learnings from Explored Alternatives}
Our exploration of a Conditional Random Field (CRF) model (Section 4.5.1) yielded a micro F1-score significantly lower than the GLiNER-BioMed baseline.
This outcome underscored the current advantages of large pre-trained transformer models, which benefit from extensive pre-training on biomedical corpora and require less manual feature engineering compared to traditional machine learning approaches like CRFs.
The substantial effort involved in crafting an effective feature set for the CRF, versus the out-of-the-box (after fine-tuning) capabilities of GLiNER-BioMed, highlighted the efficiency gains from using modern pretrained models.
Furthermore, attempts at more complex rule-based post-processing for GLiNER, such as regex-based corrections and POS-tag-based entity merging (Section 4.5.2), also proved challenging.
Regex patterns were often either too specific (missing many cases) or too general (introducing new errors).
Entity merging based on POS tags, while conceptually sound for addressing fragmented entities, did not consistently improve performance and sometimes led to incorrect merges.
These experiences reinforced our decision to focus on a more targeted, dictionary-based approach for the most prevalent and clearly defined error patterns, as broader rule-based systems proved difficult to generalize effectively.

\subsection{Limitations of the Current Study}
This study has several limitations.
Firstly, the dictionaries used for post-processing were manually curated based on errors observed primarily in the development set.
While this allowed for targeted corrections, it also likely contributed to the overfitting observed on the test set.
A more systematic and extensive dictionary creation process, potentially leveraging larger external biomedical ontologies or terminologies, might yield more robust results.
Secondly, our post-processing focused mainly on label correction and did not extensively address issues of span detection errors (false negatives or incorrect boundaries) from the GLiNER-BioMed model, which also contribute to overall performance.
Thirdly, the study relied on the initial entity candidates proposed by GLiNER-BioMed; if the model failed to identify a span entirely, our post-processing could not recover it.
Finally, the analysis of test set performance is inferential due to the lack of ground truth, relying on the comparison with development set trends.

\section{Future Work}
The outcomes of this study, particularly the generalization gap observed between the development and test sets, open several avenues for future research aimed at creating more robust BioNER systems for specialized domains like the gut-brain axis.

One primary direction is the enhancement of dictionary resources.
The dictionaries used in this work were manually curated based on development set errors.
Future efforts could focus on creating larger, more comprehensive dictionaries by leveraging existing biomedical ontologies (e.g., MeSH, Gene Ontology, ChEBI) or by semi-automatically expanding dictionaries using corpus statistics and word embeddings.
Furthermore, making dictionary rules more context-aware, perhaps by considering surrounding words or simple syntactic patterns before applying a correction, could reduce overcorrection and improve precision.

Improving the initial predictions from GLiNER-BioMed itself is also a key area.
While we performed fine-tuning, investigating different fine-tuning strategies, data augmentation techniques for the training set, or incorporating a more diverse range of biomedical texts during fine-tuning might enhance its base performance and reduce the types of errors that necessitate aggressive post-processing.

While our initial dictionary-based post-processing did not generalize as hoped, the concept of model combination or ensembling remains a viable path since simpler methods seem to reach a plateau.
Lightweight strategies could be explored, for instance, by identifying specific entity types or contexts where an alternative model (like the CRF, despite its lower overall score) demonstrably performed better than GLiNER-BioMed and selectively using its predictions.

Finally, integrating entity linking to broader knowledge bases as a post-processing or validation step could provide a more robust mechanism for both correcting labels and normalizing entities to knowledge base identifiers.
This would not only improve accuracy but also enhance the utility of the extracted entities for downstream knowledge graph construction and semantic analysis.

\section{Conclusions}
This paper presented an investigation into enhancing Biomedical Named Entity Recognition (BioNER) for the GutBrainIE CLEF 2025 Subtask 6.1, focusing on the GLiNER-BioMed model coupled with a dictionary-based post-processing strategy.
Our primary contributions include the fine-tuning of GLiNER-BioMed on the challenge's diverse training data (Platinum, Gold, and Silver collections) and the development of targeted post-processing rules to address common error patterns, particularly gene/chemical misclassifications.

Our key findings highlight a common challenge in applied machine learning: strategies that yield improvements on a development set may not always generalize to unseen test data.
The dictionary-based post-processing significantly improved the micro F1-score on our development set from a baseline of 0.79 to 0.83, primarily by correcting label inaccuracies. However, on the official blind test set, the system achieved a micro F1-score of 0.77. This outcome suggests that our system likely overfit and did not adapt well to the nuances of the test set, performing slightly below what might have been expected on unseen data.
The exploration of alternative methods, such as a CRF-based system and other rule-based post-processing techniques, further underscored the complexities of BioNER and the current advantages of large pre-trained models like GLiNER-BioMed.

In conclusion, while pre-trained models offer a powerful starting point for BioNER, domain-specific refinement through techniques like dictionary-based post-processing shows potential.
However, extra care must be taken to ensure these refinements are robust and generalizable.
This study contributes to the understanding of applying and adapting state-of-the-art NER models to specialized biomedical tasks like those in the GutBrainIE challenge and emphasizes the critical importance of strategies that mitigate overfitting and promote broader applicability.

\section*{Acknowledgements}

We thank the Data Science at Georgia Tech (DS@GT) CLEF competition group for their support.

\section*{Declaration on Generative AI}

During the preparation of this work, the author(s) used ChatGPT in order to: Polish sentences for clarity, formatting assistance, grammar and spelling check. After using these tool(s)/service(s), the author(s) reviewed and edited the content as needed and take(s) full responsibility for the publication’s content.

\bibliography{main}
\end{document}